\documentclass[10pt,twocolumn,letterpaper]{article}

\usepackage{wacv}
\usepackage{times}
\usepackage{epsfig}
\usepackage{graphicx}
\usepackage{amsmath}
\usepackage{amssymb}

\usepackage{booktabs}
\usepackage{algpseudocode}
\usepackage{algorithm}
\usepackage{mathtools}
\usepackage{color}
\usepackage{multirow,graphicx,paralist}
\usepackage{ colortbl}
\definecolor{Gray}{gray}{0.85}
\definecolor{LightCyan}{rgb}{0.88,1,1}

\DeclarePairedDelimiter\floor{\lfloor}{\rfloor}

\newcommand\Tstrut{\rule{0pt}{2.6ex}}

\usepackage[pagebackref=true,breaklinks=true,letterpaper=true,colorlinks,bookmarks=false]{hyperref}

\wacvfinalcopy 

\def\wacvPaperID{108} 

\ifwacvfinal\fi
\begin{document}

\title{Recurrent Iterative Gating Networks for Semantic Segmentation}

\author{Rezaul Karim$^1$, Md Amirul Islam$^2$, Neil D. B. Bruce$^{2,3}$ \\
$^1$University of Manitoba, $^2$Ryerson University, $^3$Vector Institute\\
{\tt\small karimr@cs.umanitoba.ca, amirul@scs.ryerson.ca, bruce@scs.ryerson.ca}
}

\maketitle
\ifwacvfinal\thispagestyle{empty}\fi

\begin{abstract}
In this paper, we present an approach for Recurrent Iterative Gating called RIGNet. The core elements of RIGNet involve recurrent connections that control the flow of information in neural networks in a top-down manner, and different variants on the core structure are considered. The iterative nature of this mechanism allows for gating to spread in both spatial extent and feature space. This is revealed to be a powerful mechanism with broad compatibility with common existing networks. Analysis shows how gating interacts with different network characteristics, and we also show that more shallow networks with gating may be made to perform better than much deeper networks that do not include RIGNet modules.
\end{abstract}

\section{Introduction}\label{sec:intro}

The problem of semantic segmentation and other pixel-wise labeling problems~\cite{long15_cvpr,chen15_iclr,zhao2017pyramid,islam2017label,badrinarayanan15_arxiv,noh15_iccv}  has been studied in detail, and is important in image understanding and for various applications including autonomous driving. Moreover, many successful solutions have been proposed based on deep neural networks~\cite{krizhevsky12_nips,simonyan15_iclr,szegedy15_cvpr,he2016deep,Huang_2017_CVPR}. While progress continues to be made, this progress has also been accompanied by increasingly deep and complex models. While increasing the depth of the network may allow for strong inferences to be made, this also risks poor spatial granularity in labeling when pooling occurs and in some sense also acts as a complex template matching process. Given a deep model, the demand of making dense predictions for all pixels at once makes the problem more challenging insofar as any relational reasoning must be handled in a single feedforward pass. There is good reason to believe that more parsimonious solutions might be presented from the careful guidance of how the image is interpreted from early layers to deep layers that encode more complex features. Evidence of this comes from both examples of existing neural networks that consider such principles~\cite{gazzaley2012top,pinheiro2014recurrent,long15_cvpr,Islam_2017_CVPR,refinenet,islam2017label,Islam2018arxiv,casanova2018iterative,carreira2016human,zamir2017feedback}, and also the very significant role that recurrence, gain control and gating play in biological vision systems as a mechanisms for task, context or input dependent adaptation.

In particular, gating is one mechanism capable of controlling information flow with the possibility of filtering out ambiguous interpretations of a scene. The precise mechanism by which this can be accomplished may take many different forms. There are at least two principal mechanisms of importance in considering the role of recurrent gating:
\begin{figure}
	\includegraphics[width=0.48\textwidth]{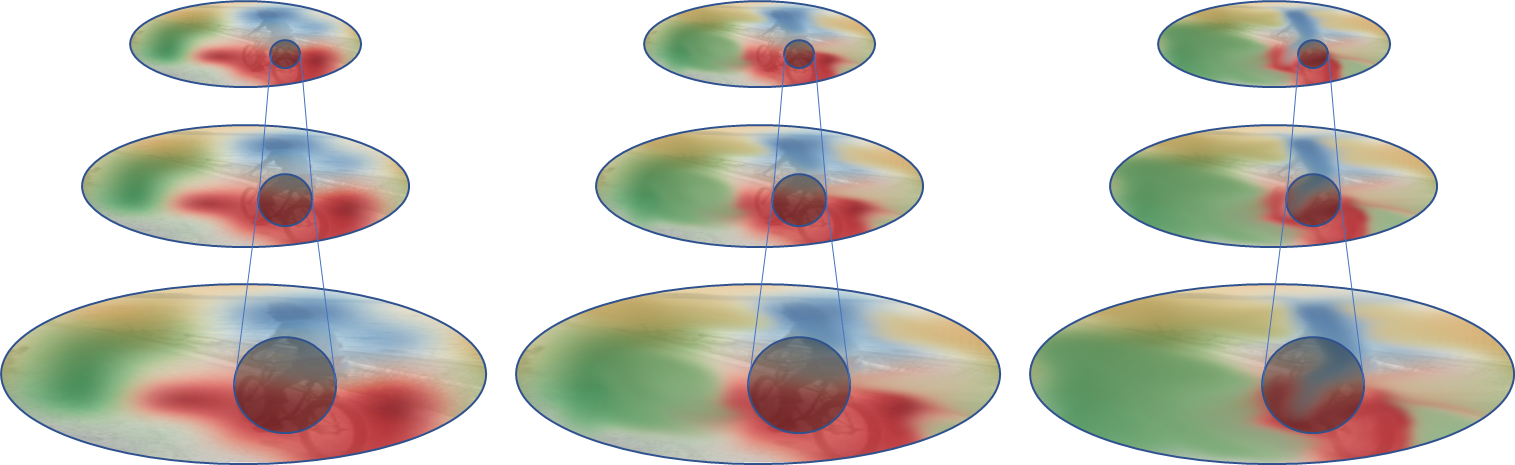}
	\caption{\textbf{A recurrent iterative gating based model}. A conceptual illustration of how higher layers of the network influence lower layers by gating information that flows forward. When applied iteratively (left to right), this results in belief propagation for features in ascending layers, that propagates over iterations both spatially and in feature space.}
	\label{fig:intro}
\end{figure}
\begin{enumerate}
\item To ensure that activation among earlier layers is consistent with the higher-level interpretation reached by later layers. For example, if a deep layer signals with high confidence that a face is present, features that carry more importance for faces, or carry more discriminative value relevant to faces should be passed forward. More importantly, activation that arises from patterns at the location of the face that are not related to faces should be suppressed.
\item In the case that local features in a scene are diagnostic of semantic category, there should be a mechanism for this local diagnosticity to propagate outwards both in features carried forward and to envelop a larger spatial extent (see Fig.~\ref{fig:intro}). That is, belief in a semantic concept may propagate spatially by virtue of the recurrent gating mechanism.
\end{enumerate}

To this end, we propose an iterative gating mechanism wherein deeper features signal to earlier features what information should be passed forward. Moreover, some networks are marked by spatial pooling, the spatial extent of activation observed very deep in the network has the capacity to propagate to a larger spatial region among earlier layers. When this is done iteratively, there is no limit to the spatial extent over which strongly diagnostic features can be matched to less discriminative and more ambiguous regions of the scene allowing for a consensus on the most likely labeling to be determined through gating and iteration.

It is also the case that one might expect that more efficient handling of the \emph{traffic} that flows through the network may allow for more parsimonious network architectures to handle challenging problems.

In this paper we present one such mechanism, a \emph{Recurrent Iterative Gating Network} RIGNet. The results presented in the paper consider how gating mechanisms may interact with different network properties including depth of network, spatial pooling, dilated filters and other characteristics. Through this presentation we show that a wide range of different network architectures reveal better performance through the use of RIG. Moreover, we show that simpler versions of networks (e.g. ResNet-50) can be as capable as much deeper networks (e.g. ResNet-101) when modified to form a RIG-Net and show that this result is true for several different networks.

This has important implications for semantic segmentation, but also any dense image labeling problem and also in how the nature of recurrent processing in general is viewed. In short, this presents a solution for improving the performance of any network by adding a simple canonical recurrent gating mechanism. 
\section{Related Work}\label{sec:background}
Feed-forward neural networks have shown tremendous success in various visual recognition tasks. e.g. image classification~\cite{simonyan15_iclr,he2016deep,Huang_2017_CVPR}, semantic segmentation~\cite{long15_cvpr,chen15_iclr,zhao2017pyramid,islam2017label,badrinarayanan15_arxiv,noh15_iccv}. Recent approaches mostly focus on generating a discriminative feature representation by increasing the depth of the network. There has also been prior research in machine learning~\cite{pinheiro2014recurrent,byeon2015scene,veit2016residual,shuai2017scene,zamir2017feedback,carreira2016human, casanova2018iterative} that incorporates feedback in the learning process. In this paper, we carefully examine the essence of feedback based learning, including the role of iterative recurrence and provide an overview of some recent work that falls into different categories in the role of feedback.

Traditional feed-forward networks, e.g., AlexNet~\cite{krizhevsky12_nips}, VGG~\cite{simonyan15_iclr} do not employ any recurrence or feedback mechanism inside the network and carry a large number of parameters. Recent successful methods e.g. ResNet~\cite{he2016deep}, GoogleNet~\cite{szegedy2015going}, and Highway Networks~\cite{srivastava2015highway} have structure connected to earlier layers in the feed-forward network in the form of residual connections. While the deeper networks with skip connections counter the problem of degradation due to depth, networks with residual connections address this with the identity shortcut. Moreover, several recent methods have shown the superiority of feedback based approaches~\cite{zamir2017feedback,carreira2016human,belagiannis2016recurrent,li2016iterative,liang2015convolutional, li2018learning,pinheiro2014recurrent} for particular tasks of interest. While most approaches focus on directly correcting the initial prediction iteratively, we instead propagate information backward through the network to learn a compact representation that is also strong in its predictions and plays an implicit role in correcting the final prediction made.

Inference in human and other biological vision systems includes both short range and long range recurrent connections\cite{gilbert2013top} where feed-forward paths work in concert with recurrent feedback to provide pre-attentive and attentive modes of vision~\cite{lamme2000distinct}. Some approaches~\cite{byeon2015scene,peng2016geometric,peng2016geometric,huang2016scene} consider recurrent neural networks (e.g. LSTM~\cite{srivastava2015highway}) for the task of semantic segmentation or scene labeling. 

Related to our proposed approach is the idea of learning in an iterative manner in a feed-forward network with feedback modulation. Recently proposed Feedback Networks~\cite{zamir2017feedback} present feedback based learning for the purpose of hierarchical taxonomy learning in image classification. 
IEF~\cite{li2016iterative} proposed a feedback based pipeline that corrects the initial prediction iteratively based on the feedback from each timestep. RCNN~\cite{pinheiro2014recurrent} introduced a recurrent convolutional neural network for scene labeling where the underlying CNN takes an input image and the label predictions from the previous iteration. 

Architecturally, our work is closest to the iterative feedback modulation strategy of recurrent feedback neural networks which are unrolled for several iterations with successive feature refinement. 
Conceptually, our work is quite different than prior works in the way that feedback is propagated through deep convolutional neural networks for feature refinement, and we provide detailed analysis and support of this approach that includes intuitive and empirical grounding. In summary, our proposed approach can be thought of as a novel formulation of feedback based recurrent design on deep convolutional neural networks that can emulate attentive vision to facilitate top down attention (in space and feature-space) and improve spatial and semantic context for inference. This implies an approach that has a high degree of generality and desirable characteristics which we relate to characteristics of different network architectures in the sections that follow. 
	\section{Recurrent Iterative Gating Network}\label{sec:approach}
In this section, we present the recurrent iterative gating mechanism with different recurrent unrolling mechanisms. We also highlight some theory towards the logical explanation for the recurrence gating module and several core advantages of different unrolling mechanisms. Finally, we propose our top-down feedback based \emph{Recurrent Iterative Gating Network}, the RIGNet for semantic segmentation.
\begin{figure}
	\begin{center}
		\includegraphics[width=0.42\textwidth]{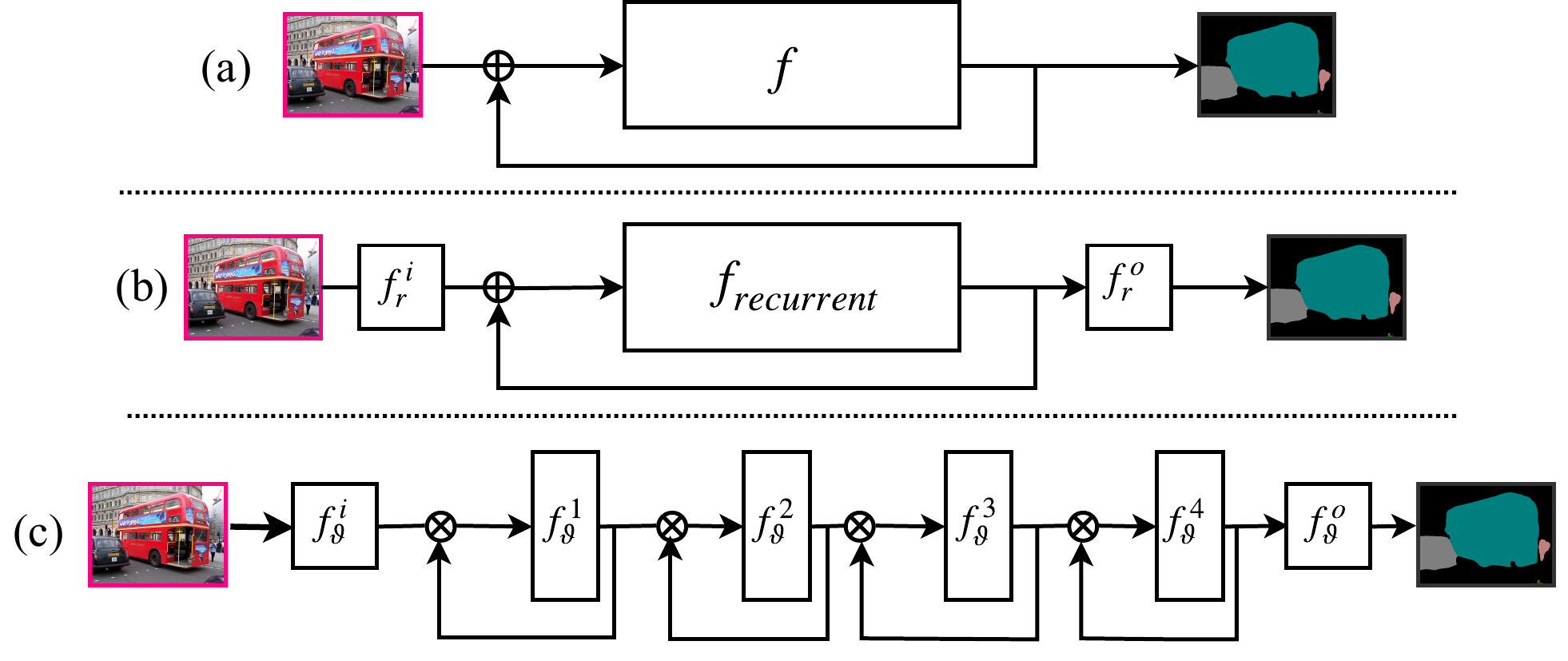}
		\caption{Illustration of (a) traditional iterative feed-forward network (b) iterative recurrent network where the hidden state serves as a feedback gate and (c) our recurrent iterative gating network for semantic segmentation.
			In contrast to previous works, our framework involves recurrent iterative gating that control the flow of information passed forward in a top-down manner.  }
		\label{fig:feedback_mechanism}
	\end{center}
	\vspace{-0.6cm}
\end{figure}
\subsection{Overview of RIGNet Formulation}
\begin{figure*}[t]
	\begin{center}
		\includegraphics[width=0.67\textwidth]{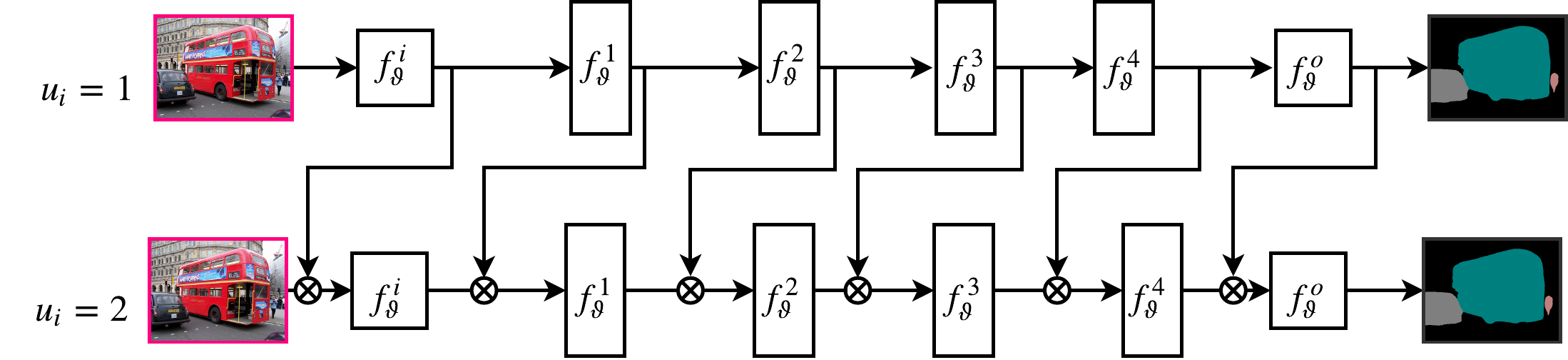}
		\caption{Our proposed network when unrolled in different time-steps ($u_i=1,2..$). The difference with the traditional approach is that the loop connections feed into layers that share parameters with previous layers in addition to gating being controlled top-down. Consider iteration $u_i = 1$ where the feed-forward network receives the input image and predicts the initial output which is gated with the corresponding iterative gating module in iteration $u_i=2$.    }
		\label{fig:feedback_unrolled}
	\end{center}
	\vspace{-0.5cm}
\end{figure*}
The process involved in iterative feedback based approaches has few key elements: (1) Output from some layers/blocks in the network is fed-back to earlier layers through a gating module where the simplest gate may be an identity transform or skip connection (2) The feedback is combined with a representation at an earlier layer through concatenation/multiplication/addition to generate the input for next iteration and (3) The final output is generated at the end of the last iteration. The input image undergoes shared convolutional stages repeatedly to predict labels at each step. The premise behind this is that iterating the feed-forward modules (most cases RNN) a few times produces a different final output at the last iteration. However, there is no backward interaction involving the feed-forward modules. In the proposed RIGNet architecture, we take the output of each block of layers as feedback, modulating the signal that forms the input of that block. The outcome of our recurrent feedback based approach is seen to be beneficial in few respects: (a) modulate the initial input with the feedback signal to emulate attentive vision (b) a compact representation that can compete performance-wise with deeper architectures through unrolling for more iteration (c) this allows a common architecture suitable for a wide range of devices with different computational throughput by varying the number of iterations (d) this introduces a hierarchical structure that leads to an implicit coarse-to-fine representations to improve spatial and semantic context of the inference. Overall, this allows for refinement of feature specific activations, and confidence for categories separated by space to propagate over the image.


An illustration of the RIGNet architecture is shown in Fig.~\ref{fig:feedback_mechanism}. Unlike existing work, we integrate iterative feedback modules inside the feed-forward network which can be seen as rerouting the captured information based on information that flows in a backward direction.  At a high-level, RIGNet mimics the cyclical structure of the human brain which is created by integrating \emph{iteration} in the feedback modules of a feed-forward network.
\subsection{Iterative Gating Mechanism}\label{sec:ifm}
In this section, we share the details of our proposed recurrent iterative gating/feedback mechanism which is based on stacking feedback modules in each stage of the feed-forward network. Recent works on semantic segmentation that have shown success typically share common strategies involving dilation~\cite{chen2016deeplab,YuKoltun2016,Huang_2017_CVPR}, encoder-decoder structure~\cite{noh15_iccv,islam2017label,badrinarayanan15_arxiv}, and coarse-to-fine refinement~\cite{islam2017label,pinheiro2016learning,casanova2018iterative,refinenet} to balance semantic context and fine details by recovering per-pixel categorization. Most of these approaches are increasingly precise in their performance but introduce additional model complexity. Our main objective is to demonstrate a compact representation of complex deep networks which can achieve similar performance and is agnostic to the network it is paired with. To accomplish this, we propose a network with several iterative feedback modules (shown as a feedback gate in Fig.~\ref{fig:feedback_mechanism}) similar to recurrent neural networks. More specifically, we apply recurrent top-down feedback in a block-wise manner rather than layer-wise~\cite{liang2015recurrent, liang2015convolutional} or over the full network~\cite{pinheiro2014recurrent}. We argue that formulating top-down feedback layer-wise similar to~\cite{liang2015recurrent, liang2015convolutional} suffers from few major drawbacks: (1) adds a huge overhead in number of parameters (2) limiting for transfer learning (3) the output of a single deeper layer may not contain sufficiently rich semantic information for feedback. Moreover, the recurrence over the whole network~\cite{pinheiro2014recurrent} is also unable to leverage semantic and spatial contextual information through feedback to refine the previous layers. In this context, the behaviour of filters remains fixed and only varies based on the current label hypotheses rather than any internal feature representations. In contrast, in RIGNet feedback is taken from the output of a block of convolution layers resulting in a larger effective field of view and more abstraction. Also, this mechanism allows the lower-level layers to be influenced by the weight/activation of higher-level features resulting in refined weight/activation in the earlier layers that refines information passed forward and importantly also allows for spatial propagation through internal feature representations throughout the network. The mechanism is further defined by a parameter ($u_i$) denoted as the \textit{unroll iteration} of the network that determines the number of times the feedback loop will be instantiated to predict the final output. The unrolling parameter ($u_i$) can be viewed the same as the unrolling steps in RNNs. Fig.~\ref{fig:feedback_unrolled} shows the unrolling effect on the network presented in Fig.~\ref{fig:feedback_mechanism} (c).

The final output is generated with multiple iterations over the network where the number of iterations is determined by the unroll iterations ($u_i$). In the very first iteration, all feedback gates remain disconnected from the network (i.e. there is no gating without prior knowledge of the image among subsequent layers). In subsequent iterations, the output of the previous iteration is subject to modulation by the feedback gate at every layer in the most extreme case all the way from the first layer to the deepest layers. This operation repeats for all the stages of the network to obtain a prediction for the current iteration. The reverse information flow towards the input all the way from output layers allows the earlier layers to be implicitly subject to adjusted parameters at the time of inference to provide guidance to remove ambiguity that may arise anywhere within the network. Similarly, the loopy structure inside the feedback modules allows a shallower network to be influenced by a rich feature representation. The iterative nature allows for stage-wise refinement and spatial propagation of refinement. This implies that much simpler networks can perform more powerful inference given that their behaviour is not fixed and can be modulated by recurrent feedback.
\subsection{Unroll Mechanism}\label{sec:unroll_mechanism}
For a recurrent approach to be practical, there is a need for a specific recurrent structure (and implied unrolling mechanism) and also a constraint that the unrolling iterations should be finite. It is important to note that often only a finite number of iterations is of value given that the final output will tend to converge on a certain set of labels and show little change beyond a fixed number of iterations. Note also that when we set the unroll iteration $u_i = 1$, RIGNet becomes a purely end-to-end feed-forward neural network. We explore two different unrolling mechanisms to control the flow of information in a top-down manner.
\subsubsection{Sequential Unroll}\label{sec:sequential_unroll}
\begin{figure}[h]
	\vspace{-0.2cm}
	\begin{center}
		\includegraphics[width=0.45\textwidth]{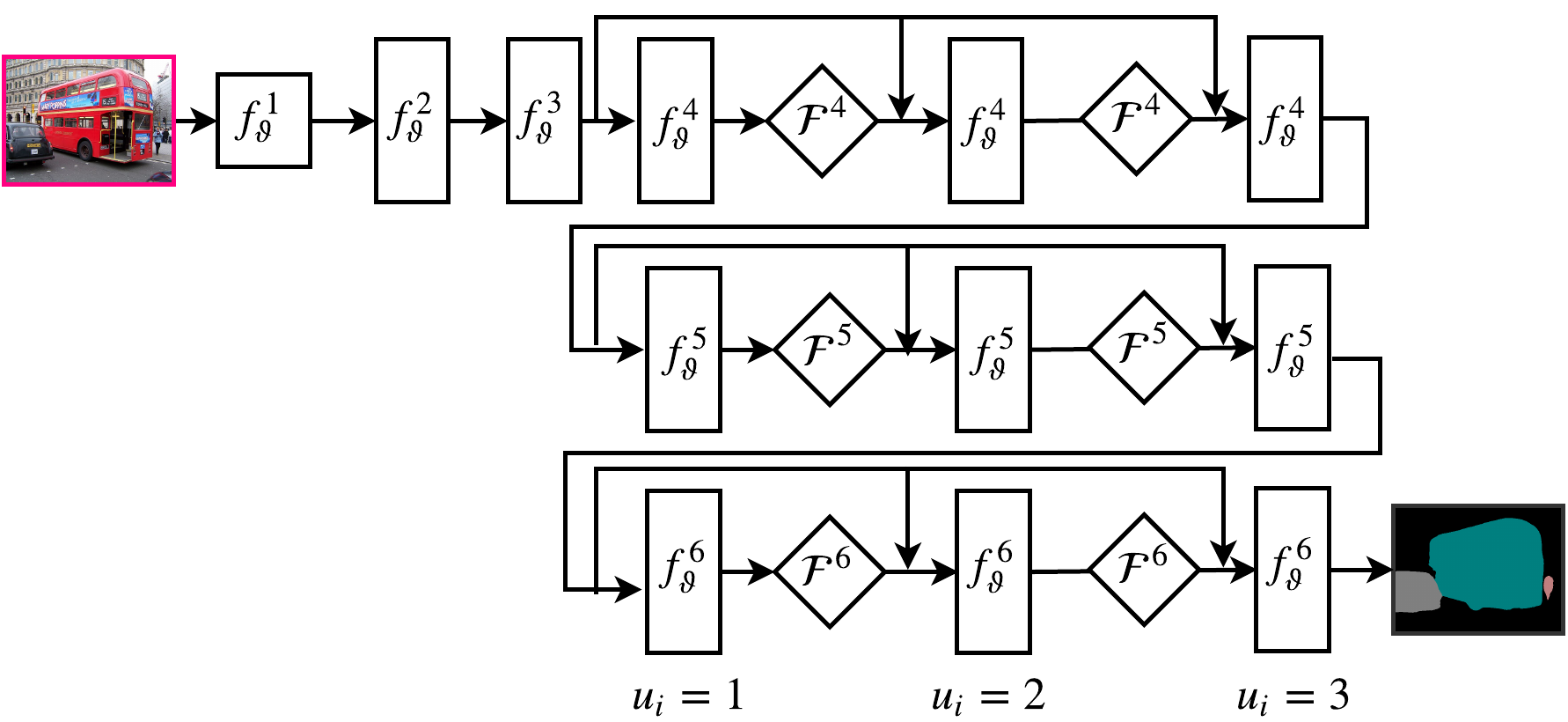}
		\caption{Illustration of RIGNet with sequential unroll for three iteration in last three feed-forward blocks. Note that $f_\vartheta^i$ refers to a feed-forward block and $\mathcal{F}^i$ denotes the recurrent feedback gate.}
		\label{fig:sequential_unroll}
	\end{center}
	\vspace{-0.8cm}
\end{figure}

Sequential unrolling involves the recurrence within block tied to a particular iteration, with the order of recurrence following subsequent blocks all involved in a single feedforward pass. Activations are forwarded to the next recurrent block when iterations in all previous network blocks within a single recurrent iteration are finished. Fig.~\ref{fig:sequential_unroll} depicts the sequential unrolling mechanism when the network is unrolled for three iterations ($u_i = 3$). Conceptually, the sequential unroll mechanism increases the effective depth of the unrolled network by a multiplicative factor. For example, given a network with $l_t$ layers in the RIGNet formulation, $l_f$ of $l_t$ layers remain similar whereas we introduce $l_r$ layers within the blockwise recurrent feedback stage. So, the feed-forward network has effective depth similar to the original depth of the network ( $l_e=l_f+ l_r$) however the RIGNet with unroll iteration $u_i$ has effective depth of $l_e=l_f+l_r*u_i$.
\subsubsection{Parallel Unroll}\label{sec:parallel_unroll}
\begin{figure}[t]
	\begin{center}
		\includegraphics[width=0.45\textwidth]{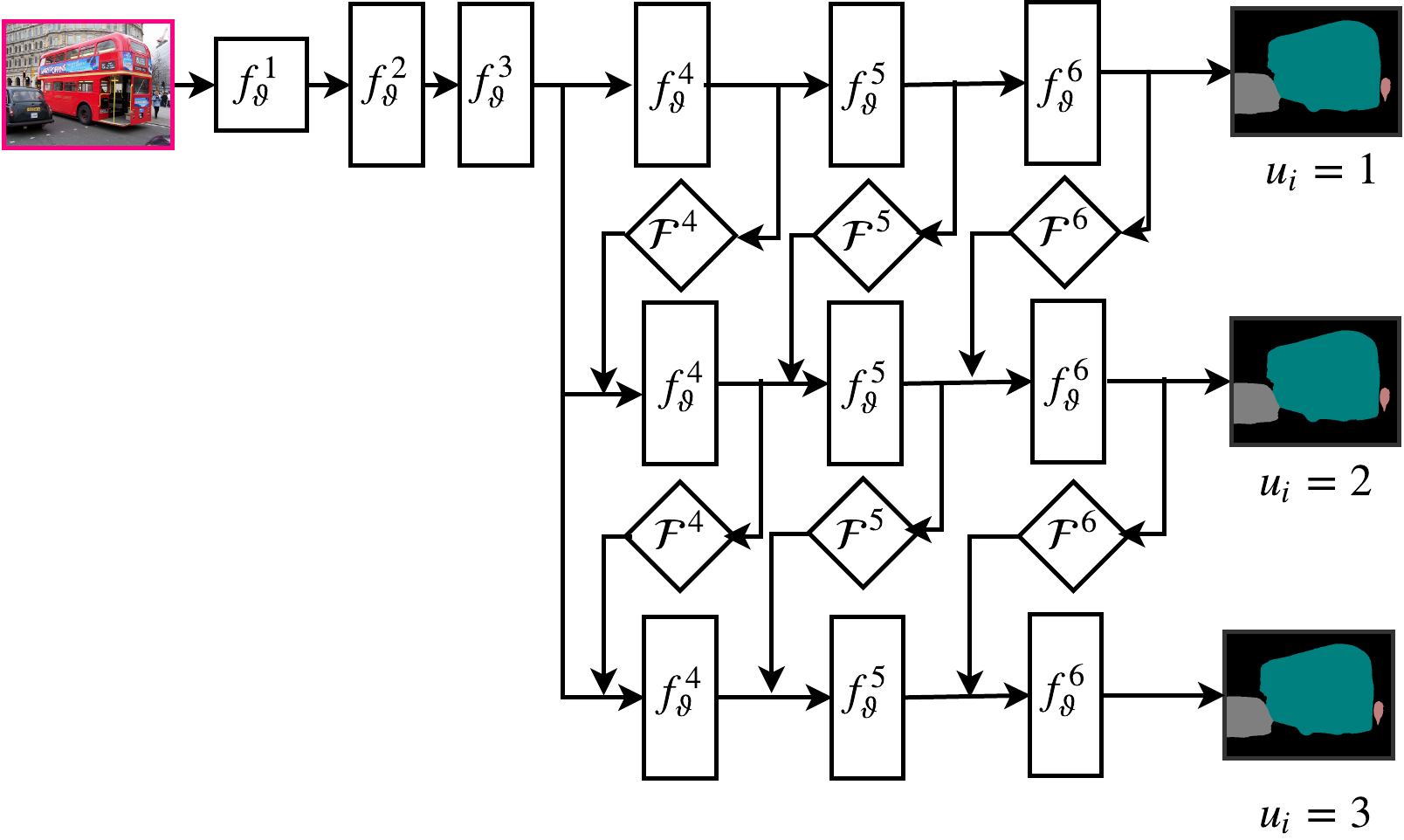}
		\caption{Illustration of RIGNet with parallel unroll in the last three feed-forward blocks for three iterations.}
		\label{fig:parallel_unroll}
	\end{center}
	\vspace{-0.8cm}
\end{figure}

In the parallel unrolling mechanism, the network initially gathers the final representation (activations) of the first iteration and then recurrence proceeds by way of feedback from the first block to the last (deep $\rightarrow$ shallow). The formulation of obtaining the final representation in subsequent iterations is similar to the first iteration. Our proposed RIGNet in Fig.~\ref{fig:feedback_unrolled} is a feed-forward network with a parallel unrolling mechanism.  Fig.~\ref{fig:parallel_unroll} illustrates the parallel unrolling mechanism where a RIGNet with feedback in last three blocks is unrolled for three iterations. The parallel unrolling mechanism has less effective depth compared to the sequential one since it increases the effective depth multiplicatively only for the last feedback block. To continue with the example in Sec.~\ref{sec:sequential_unroll}, assume the last feedback block has $l_{rk}$ layers among $l_r$ whereas the remaining blocks with feedback have $l_{rj}=l_r-l_{rk}$ layers.
In this case the effective depth of the RIGNet with parallel unrolling is $l_e=l_f+l_{rj}+l_{rk}*u_i$. This depth analysis between sequential and parallel unrolling mechanisms reveals the impact of increasing the effective depth on performance improvement.
This analysis also sheds light on the hypothesis that an effective \textit{depth increase} helps to improve the overall performance as opposed to the role of long range semantic context (across layers and spatially) through feedback bringing improvement. Additionally, parallel recurrence brings another advantage of generating a early prediction (coarse representation) at the end of each iteration which can be used to facilitate taxonomy learning where some coarse grained predictions can be made at initial iterations (like vehicles) and then fine predictions in final iterations (like cars/bus). This can be done by backpropagation of loss defined by coarse predictions in initial iteration similar to the concept explored in~\cite{zamir2017feedback} for classification task. Since we are principally interested in the final prediction, we only backpropagate loss once at the end of final iteration, but it should be noted that the generality of the RIGNet structure presents a wide range of directions that may be explored further.
\subsubsection{Parallel Unroll with Multi Range Feedback}\label{sec:pu_mrf}
In addition to our basic feedback model (RIGNet), we also extend the concept of short range feedback to long range feedback. Our multi range feedback is a best fit for the parallel recurrence mechanism as it requires additional feedback from not only subsequent blocks but also deeper blocks. The concept of multi range feedback is more analogous to biological vision systems and has been considered to a limited degree in the context on image classification~\cite{nayebi2018task}.

\subsection{Iterative Gating Module Formulation}\label{sec:ifg}
In this section we present the formulation for both our basic feedback gate and multi range feedback gate.

\subsubsection{Basic Feedback Gate} 
Each feedback gate module takes the output of the forward block (next stage feature map) $f_\vartheta^{i+1}$ as input and learns to pass information relevant to gating backwards through the following sequence of operations: First we apply an average pooling (resultant feature map $f_\vartheta^{i^\prime+1}$) followed by a $3\times3$ convolution to obtain a feature map $f_b^{i^\prime+1}$ which is capable of carrying context with a larger field of view. We then apply a sigmoid operation followed by bilinear upsampling to produce a feature map whose spatial resolution is the same as the input to the feedback gate. The resultant feature map provides the input to the feedback gate. The $i^{th}$ stage feedback gate $\mathcal{F}^{i+1}$ combines the inputs through an element-wise product resulting in a modulated feature map $f_r^{i}$.

\noindent We can summarize these operations as follows: 	
\begin{gather}
f_b^{i^\prime+1} = \mathcal{S}(\mathcal{C}_{3\times3}(\mathcal{A}_p(f_\vartheta^{i+1});\Theta), \hspace{0.2cm} f_b^{i^\prime+1} = \xi(f_b^{i^\prime+1}) \nonumber \\
f_r^{i} = f_b^{i^\prime+1} \otimes f_\vartheta^{i-1}
\end{gather}
where $\mathcal{A}_p$ represents an average pooling operation and ($\Theta$) denotes the parameters of the convolution $\mathcal{C}$. $\xi$ refers to upsampling operation.
\subsubsection{Multi Range Feedback Gate}
In a multi range feedback gate, the long range feedback is taken from the output of the last block. For example, we denote the output from the last block as $f_c^i$ at iteration $i$. The long range feedback $f_c^i$ is first passed through convolution and an interpolation layer resulting in a feature map of $f_c^{i^\prime}$ with a number of channels and spatial dimensions matching the short range feedback $f_\vartheta^{i+1}$ (see Eq.1). We then concatenate $f_c^i$ and $f_\vartheta^{i+1}$ to obtain a feature map $f_c^{i^\prime}$. The rest of the operations of the feedback gate are similar to the basic feedback gate with $f_c^{i^\prime}$ instead of $f_\vartheta^{i+1}$.   
\section{Experiments}\label{sec:exp}
We perform a series of experiments to evaluate the performance of RIGNet. We begin by providing implementation details of RIGNet and baseline approaches. Initially, we perform a controlled study to analyze and investigate the contribution of each component in RIGNet. Then we experiment on the object-centric PASCAL VOC 2012 dataset~\cite{everingham2015pascal}. In addition to semantic segmentation, we also explore how the recurrent iterative gating mechanism can improve scene parsing performance on the COCO-Stuff dataset~\cite{caesar2016coco}.
\subsection{Implementation Details and Baseline Networks}
Our experimental pipeline and the pre-trained models are based on the open source toolbox PyTorch~\cite{paszke2017automatic}. 
We begin with experiments using simpler networks and gradually move to more complex networks to show performance improvements related to different network architectures with the recurrent iterative gating mechanism.
First, we report evaluation performance for the vanilla ResNet baselines denoted as ResNet50-FCN, ResNet101-FCN and our corresponding proposed ResNet50-RIGNet, ResNet101-RIGNet networks. So given an input image $I\in\mathbb{R}^{h\times w\times d}$, the networks produce a feature map of size  $\floor*{\frac{h}{32}, \frac{w}{32}}$. Then we extend our experiments with a more sophisticated network architecture to examine the effectiveness of RIGNet. We choose DeepLabV2~\cite{chen2018deeplab} as our new baseline network due to it's superior performance on pixel-wise labeling tasks. DeepLabV2 uses the dilated structure to balance the semantic context and fine details, resulting in a feature map of size $\floor*{\frac{h}{8}, \frac{w}{8}}$ given an input image $I\in\mathbb{R}^{h\times w\times d}$. 

For ease of representation, we use the following notations to report numbers throughout the experiment section. Sequential unroll ($\mathbb{S}_u$), Parallel unroll ($\mathbb{P}_u$), Parallel unroll with multi-range feedback ($\mathbb{P}_u^{f}$), unroll iteration ($u_i$). $\mathbb{F}_b\ll j..i \gg$ refers to feedback used in blocks from $i$ to $j$.
\subsection{Analyzing RIGNet}
In this section, we perform ablation studies to investigate the role of the \emph{recurrent iterative gating} mechanism. We highlight a few major facts to validate the design choices: 1) the role of applying iterative gating modules in different stages 2) the length of iteration in a gating module. 3) the design choice associated with the feedback gate 4) the influence of network-wide vs block-wise feedback mechanism.
\subsubsection{Unroll Mechanism and Feedback Blocks}
To evaluate the value of applying iterative gating modules in different convolutional stages, we perform a control study where we train models by adding iterative gating modules step by step to different layers to evaluate their effect on performance. More specifically, we train a feed-forward network by adding feedback modules at the last stage only and compute mIoU for the final predictions. We repeat this operation several times until we reach the first convolutional stage to examine the importance of integrating recurrent iterative gating mechanisms at the final layer, many deep layers, or all layers. We first report mIoU for the vanilla ResNet baselines in Table~\ref{tab:voc2012_val_fcnresnet_baselines}. Table~\ref{tab:voc2012_val_fcnresnet_mechanism} shows the depth-wise performance of the ResNet50-RIGNet architecture on the PASCAL VOC 2012 validation set. From this analysis, it is clear that inclusion of iterative gating modules improves the overall performance gradually.
\begin{table}[h]
	\vspace{-0.2cm}
	\begin{center}
		\def\arraystretch{1.1}
		\resizebox{0.27\textwidth}{!}{
			\begin{tabular}{c|c|c}
				\specialrule{1.2pt}{1pt}{1pt}
				Methods& Parameters & Mean IoU  \\
				\specialrule{1.2pt}{1pt}{1pt}
				ResNet50-FCN & 32-s &59.4\\
				ResNet101-FCN & 32-s & 65.3 \\
				\specialrule{1.2pt}{1pt}{1pt}
			\end{tabular}}
			\caption{Quantitative results in terms of mIoU on PASCAL VOC 2012 validation set for ResNet-FCN based baselines.}
			\label{tab:voc2012_val_fcnresnet_baselines}
		\end{center}
		\vspace{-0.8cm}
	\end{table}
\begin{table}[h]
	\begin{center}
		\def\arraystretch{1.1}
		\resizebox{0.48\textwidth}{!}{
			\begin{tabular}{c|c|c|c|c|c|c}
				\specialrule{1.2pt}{1pt}{1pt}
				\multirow{ 2}{*}{Methods}& \multicolumn{6}{c}{Feedback Blocks}\\
				& $\ll$6$\gg$  & $\ll$6..5$\gg$ & $\ll$6..4$\gg$ & $\ll$6..3$\gg$&$\ll$6..2$\gg$&$\ll$6..1$\gg$   \\
				\specialrule{1.2pt}{1pt}{1pt}
				$\mathbb{S}_u$&61.6 & 65.1& \cellcolor{Gray} \color{red}\textbf{65.4} &65.2 &65.3 &65.3 \\
				$\mathbb{P}_u$&62.3 & 64.8 & \cellcolor{Gray} \color{red}\textbf{65.2} & 64.9&64.8 &65.1 \\
				$\mathbb{P}_u^{f}$&61.6 &66.3 & 66.7& 66.8&67.1 &\cellcolor{Gray} \color{red}\textbf{67.2} \\
				\specialrule{1.2pt}{1pt}{1pt}
			\end{tabular}}
			\caption{Comparison of mIoU for different variants of our proposed ResNet50-RIGNet architecture.} 
			\label{tab:voc2012_val_fcnresnet_mechanism}
		\end{center}
		\vspace{-0.8cm}
\end{table}

\subsubsection{Unroll Iteration}
To justify the significance of an iterative solution, we examine the extent of the recurrent gating module in terms of iterations. Table~\ref{tab:voc2012_val_fcnresnet_ui} shows the experimental results of the iterative gating module per the discussion. We keep the best performing combination for the three different scenarios.
\begin{table}[h]
	\begin{center}
		\vspace{-0.1cm}
		\def\arraystretch{1.1}
		\resizebox{0.49\textwidth}{!}{
			\begin{tabular}{c|c|c|c}
				\specialrule{1.2pt}{1pt}{1pt}
				Iter.& RIGNet ($\mathbb{S}_u$) $\ll$6..4$\gg$ & RIGNet ($\mathbb{P}_u$) $\ll$6..4$\gg$  & RIGNet ($\mathbb{P}_u^{f}$) $\ll$6..1$\gg$ \\
				\specialrule{1.2pt}{1pt}{1pt}
				2 &65.4&65.2&67.2\\
				4&68.3&68.3&68.5\\
				6&\cellcolor{Gray} \color{red}\textbf{68.8}&\cellcolor{Gray} \color{red}\textbf{68.9}& \cellcolor{Gray} \color{red}\textbf{69.5}\\
				\specialrule{1.2pt}{1pt}{1pt}
			\end{tabular}}
			\caption{Comparison of mIoU for varying number of unroll iterations with ResNet50-RIGNet variants on PASCAL VOC 2012.}
			\label{tab:voc2012_val_fcnresnet_ui}
		\end{center}
		\vspace{-0.4cm}
\end{table}

For this analysis, we compare evaluation performance for \textit{unroll iter 2, unroll iter 4, and unroll iter 6}. We observe that overall performance progressively improves with each successive stage of iteration. We empirically found this observation to be valid across datasets and different network architectures. Interestingly, ResNet50-RIGNet with unroll iteration 3 outperforms the ResNet101-FCN which further validates the impact of increasing the number of iterations in recurrent gating.
\subsubsection{Feedback Gate Design Choices} Additionally, concerning the design choice of feedback gate, we try a variety of alternative design choices and report the number in Table~\ref{tab:fb_gate_design}. When we use (additive + ReLU) interaction in recurrent gating modules, ResNet50-RIGNet achieves 64.2\% mIoU on the PASCAL VOC 2012 validation set. In comparison, our proposed ResNet50-RIGNet with an (multiplicative + sigmoid) interaction in the gating modules achieves 65.2\% mIoU. Multiplicative feedback routing is demonstrably valid from a performance point of view, but also intuitive in that it provides a stronger capacity to resolve categorical ambiguity present among earlier layers in the extreme case completely inhibiting activation in an earlier layer.
\begin{table}[h]
	\begin{center}
		\def\arraystretch{1.1}
		\resizebox{0.45\textwidth}{!}{
			\begin{tabular}{c|c|c|c }
				\specialrule{1.2pt}{1pt}{1pt}
				Methods&Add + ReLU&Mul + Sigmoid&Mul + Tanh\\
				\specialrule{1.2pt}{1pt}{1pt}
				ResNet50-RIGNet & 64.2&\cellcolor{Gray} \color{red}\textbf{65.2} &65.2\\
				ResNet101-RIGNet & 67.0&\cellcolor{Gray} \color{red}\textbf{68.9 }&68.0\\
				\specialrule{1.2pt}{1pt}{1pt}
			\end{tabular}
		}
		\caption{Impact of the choice of activation function in iterative gating on PASCAL VOC 2012 validation set. }
		\label{tab:fb_gate_design}
	\end{center}
	\vspace{-0.3cm}
\end{table}

Moreover, we also investigate the impact of different pooling operations (w/o pooling, max pool, and avg pool) in the design choice of recurrent iterative gating. Table~\ref{tab:fb_gate_pooling} presents the results of alternative design choice in terms of different pooling operations.
For ResNet50-RIGNet all the three different design choices achieve similar mIoU on PASCAL VOC 2012 val set. Interestingly, the ResNet101-RIGNet with an average pooling in the recurrent gating achieves better performance compared to alternatives.
\begin{table}[h]
	\begin{center}
		\def\arraystretch{1.1}
		\resizebox{0.42\textwidth}{!}{
			\begin{tabular}{c|c|c|c }
				\specialrule{1.2pt}{1pt}{1pt}
				Methods&w/o Pooling&Max Pool&Average Pool\\
				\specialrule{1.2pt}{1pt}{1pt}
				ResNet50-RIGNet& 65.2&65.2 &\cellcolor{Gray} \color{red}\textbf{65.2}\\
				ResNet101-RIGNet& 68.3&68.3 &\cellcolor{Gray} \color{red}\textbf{68.9}\\
				\specialrule{1.2pt}{1pt}{1pt}
			\end{tabular}
		}
		\caption{Impact of choice of pooling operation in the iterative gating with $\mathbb{P}_u$$\ll$6..4$\gg$, $u_i = 2$. }
		\label{tab:fb_gate_pooling}
	\end{center}
	\vspace{-0.7cm}
\end{table}
\subsubsection{Feedback: Network-wide vs Block-wide}
In Table \ref{tab:fb_recurrence_design}, we present the results comparing different feedback routing mechanisms shown in Fig.~\ref{fig:feedback_mechanism}. Note that existing works incorporate network-wide recurrence~\cite{pinheiro2014recurrent} with a shallower base network and the results comply with our general intuition of the superiority of the gated block-wide feedback mechanism.
\begin{table}[h]
	\begin{center}
		\def\arraystretch{1.1}
		\resizebox{0.45\textwidth}{!}{
			\begin{tabular}{c|c|c }
				\specialrule{1.2pt}{1pt}{1pt}
				Methods&Recurrence Mechanism&mIoU\\
				\specialrule{1.2pt}{1pt}{1pt}
				ResNet50-FCN &feed-forward&59.4\\
				RCNN & Network-wide, similar to \cite{pinheiro2014recurrent} &59.6\\
				RCNN & Network-wide gated feedback &59.9\\
				ResNet50-RIGNet & routing similar to \cite{li2018learning} & 65.0\\
				ResNet50-RIGNet &$\mathbb{P}_u \ll$6..4$\gg$, $u_i=2$ &65.2\\
				ResNet50-RIGNet &$\mathbb{P}_u^{f} \ll$6..1$\gg$, $u_i=2$ &\cellcolor{Gray} \color{red}\textbf{67.2}\\				
				\specialrule{1.2pt}{1pt}{1pt}
			\end{tabular}
		}
		\caption{Comparison of network-wide vs block-wide recurrence based methods on PASAL VOC val. Note that we implement the ideas in \cite{pinheiro2014recurrent} with ResNet50. Also, we adapt~\cite{li2018learning} by implementing their routing mechanism attached with our basic feedback gate.  }
		\label{tab:fb_recurrence_design}
	\end{center}
	\vspace{-0.5cm}
\end{table}
\subsection{Experiments on PASCAL VOC 2012}
We evaluate the performance of our proposed recurrent iterative gating network on the PASCAL VOC 2012 dataset, one of the most commonly used semantic segmentation benchmarks. Following prior works~\cite{long15_cvpr,chen2018deeplab,Islam_2017_CVPR}, we use the augmented training set comprised of 10,581, 1449, and 1456 images in training, validation, and testing respectively. The models are trained on the augmented training set and tested on the validation set. Table~\ref{tab:voc2012_val_fcnresnet} shows results for the comparison between our proposed approach and the ResNet baselines on the validation set.
\begin{table}[h]
	\begin{center}
		\def\arraystretch{1.1}
		\resizebox{0.45\textwidth}{!}{
			\begin{tabular}{c|c|c}
				\specialrule{1.2pt}{1pt}{1pt}
				Method&Parameters &mIoU (\%)\\
				\specialrule{1.2pt}{1pt}{1pt}
				ResNet50 &-&59.4\\
				ResNet50-RIGNet& $\mathbb{P}_u$$\ll$6..4$\gg$, $u_i=6$ &\cellcolor{Gray} \color{red}\textbf{68.9}\\
				\midrule
				ResNet101 &-&65.3\\
				ResNet101-RIGNet& $\mathbb{S}_u$$\ll$6..4$\gg$, $u_i=6$&71.2\\
				ResNet101-RIGNet& $\mathbb{P}_u$$\ll$6..4$\gg$, $u_i=6$&71.4\\
				ResNet101-RIGNet& $\mathbb{P}_u^{f}\ll6..1\gg$, $u_i=6$&\cellcolor{Gray} \color{red}\textbf{71.6}\\
				\midrule
				ResNet101 (8s) &-&71.3\\
				ResNet101-RIGNet(8s)& $\mathbb{P}_u$$\ll6..4\gg$, $u_i=4$&\cellcolor{Gray} \color{red}\textbf{74.9}\\
				\midrule
				DeepLabV2&- & 74.9 \\
				DeepLabV2-RIGNet& $\mathbb{P}_u$$\ll6..4\gg$, $u_i=2$ &\cellcolor{Gray} \color{red}\textbf{75.9}\\

				\specialrule{1.2pt}{1pt}{1pt}
			\end{tabular}}
			\caption{PASCAL VOC 2012 validation set results for several baselines and RIGNet with different unroll mechanisms. }
			\label{tab:voc2012_val_fcnresnet}
		\end{center}
		\vspace{-0.3cm}
\end{table}

Note that we only report the best result for ResNet50-RIGNet in Table~\ref{tab:voc2012_val_fcnresnet} since the depth-wise results are already reported in Table~\ref{tab:voc2012_val_fcnresnet_mechanism}. It is evident that, \textit{ResNet50-RIGNet} and \textit{ResNet101-RIGNet} outperform the baselines significantly in terms of mIOU achieving 68.9\%  and 71.6\% respectively. It is worth mentioning that our proposed ResNet50-RIGNet yields mIoU better than ResNet101-FCN without any post-processing techniques providing a convincing case for the value of our iterative gating mechanism. We also experiment with ResNet101-FCN (stride 8) as a base network and achieve superior performance (71.3\% vs. 74.9\% mIoU) compared to the baseline. As shown in Table~\ref{tab:voc2012_val_fcnresnet}, \textit{DeepLabv2-RIGNet} outperforms the baseline significantly which further validates the importance of iteratively refining initial outcomes through recurrent gating modules.

\begin{figure}[h!]
	\begin{center}
		\setlength\tabcolsep{0.7pt}
		\def\arraystretch{0.5}
		\resizebox{0.48\textwidth}{!}{
			\begin{tabular}{*{5}{c }}

			    	\includegraphics[width=0.2\textwidth]{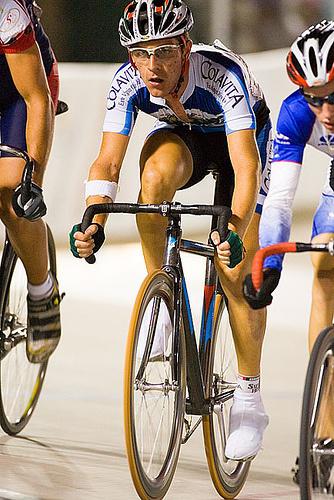}&
			    	\includegraphics[width=0.2\textwidth]{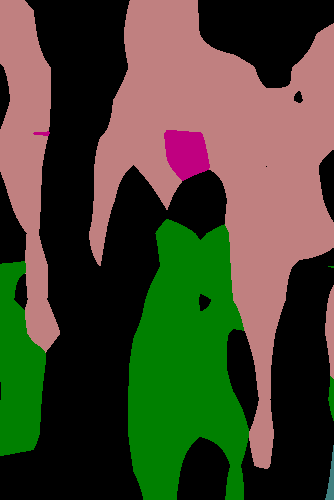}&
			    	\includegraphics[width=0.2\textwidth]{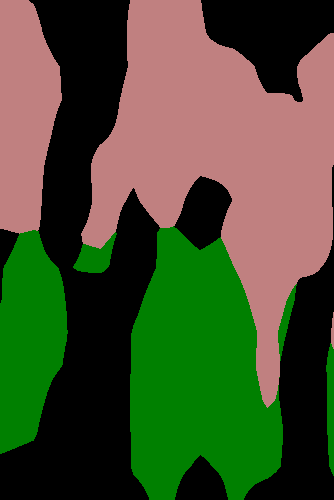}&
			    	\includegraphics[width=0.2\textwidth]{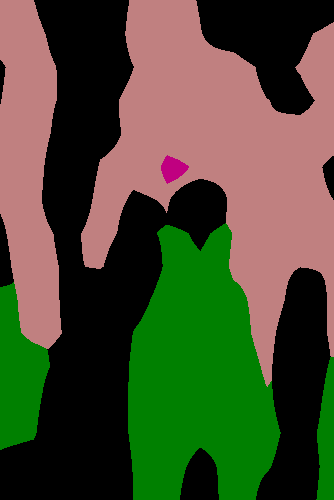} &
			    	\includegraphics[width=0.2\textwidth]{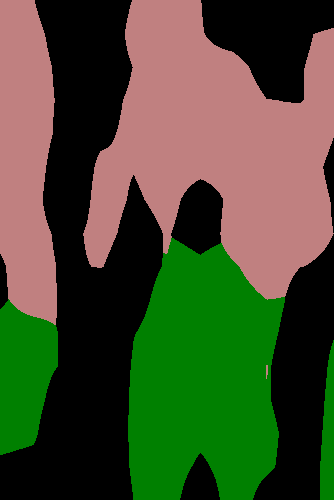}\\

			    	\includegraphics[width=0.2\textwidth]{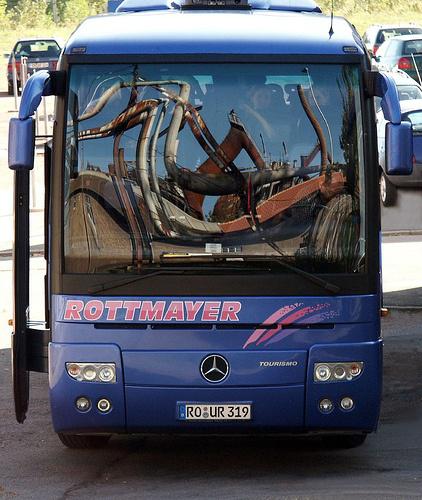}&
			    	\includegraphics[width=0.2\textwidth]{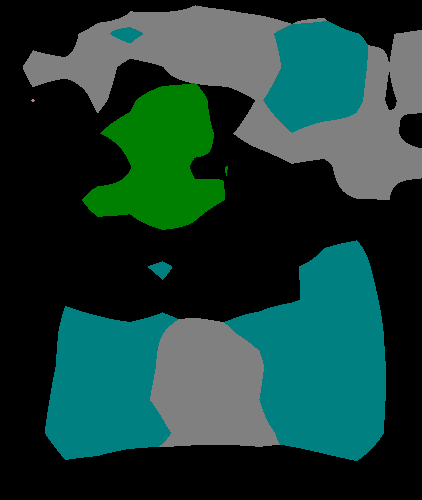}&
			    	\includegraphics[width=0.2\textwidth]{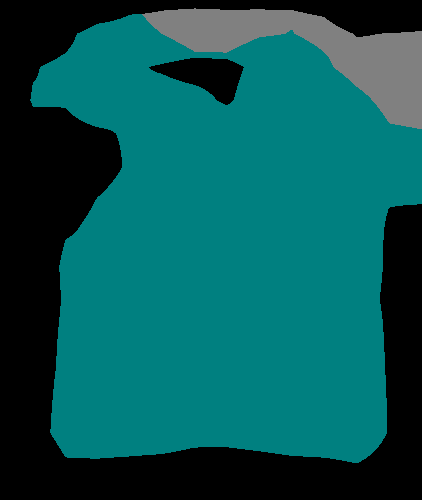}&
			    	\includegraphics[width=0.2\textwidth]{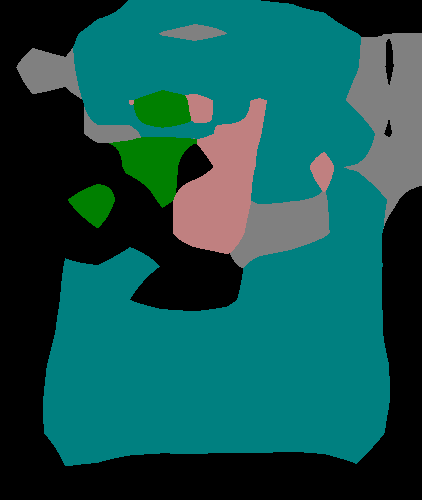} &
			    	\includegraphics[width=0.2\textwidth]{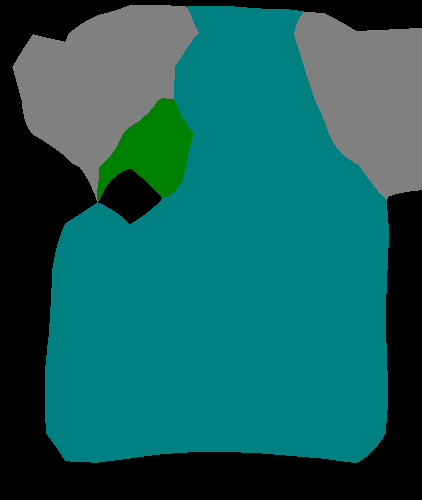}\\

				Image & ResNet50-FCN & \textbf{ResNet50-RIGNet} & ResNet101-FCN& \textbf{ResNet101-RIGNet} \Tstrut
			\end{tabular}}
			\caption{Qualitative results corresponding to the PASCAL VOC 2012 validation set for 2 iteration.}
			\label{fig:val_pascal}
		\end{center}
		\vspace{-0.5cm}
\end{figure}

Figure~\ref{fig:val_pascal} depicts a visual comparison of our approach with respect to the baselines. We can see that \textit{ResNet50-RIGNet} is capable of producing predictions superior to ResNet101-FCN, and the RIG mechanism has a powerful impact on network performance for all cases.
\subsection{Experiments on COCO-Stuff }
To further confirm the value and generality of proposed \textit{recurrent iterative gating} mechanism on scene parsing, we evaluate on the large-scale COCO-Stuff dataset. This dataset contains images of high-complexity including things and stuff. The COCO-Stuff dataset extends the COCO annotation by adding dense pixel-wise stuff annotations and provides dense semantic labels for the whole scene, which has 9,000 training images and 1,000 test images. It includes total annotation of 182 classes consisting of 91 thing classes and 91 stuff classes.
\begin{figure*}[ht]
	\begin{center}
		\setlength\tabcolsep{1.1pt}
		\def\arraystretch{0.5}
		\resizebox{0.99\textwidth}{!}{
			\begin{tabular}{*{8}{c }}
				
				\includegraphics[width=0.2\textwidth]{images/pascal/original/2007_000663.jpg}&
				\includegraphics[width=0.2\textwidth]{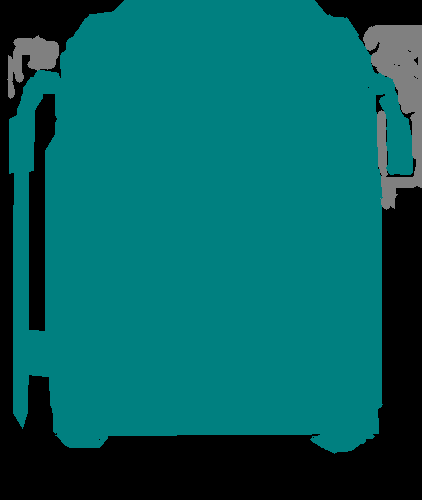}&
				\includegraphics[width=0.2\textwidth]{images/pascal/ResNet50-FCN/2007_000663.png}&
				
				\includegraphics[width=0.2\textwidth]{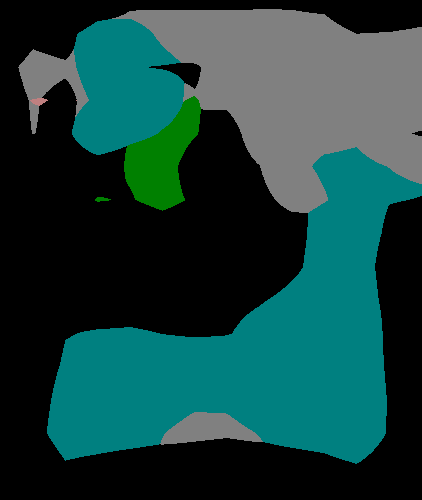}&
				\includegraphics[width=0.2\textwidth]{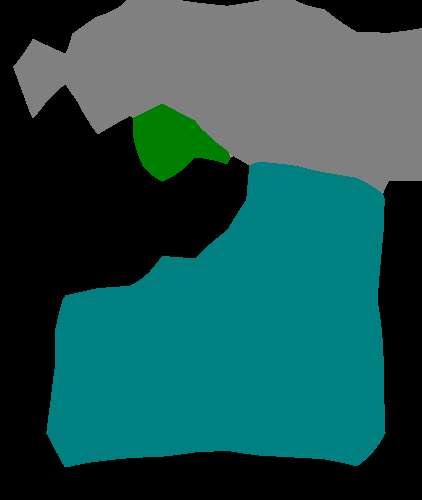} &
				\includegraphics[width=0.2\textwidth]{images/pascal/ResNet50-RIGNet/2007_000663.png} &
				\includegraphics[width=0.2\textwidth]{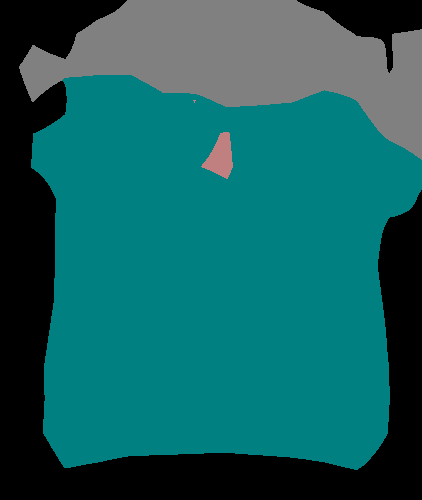}&
				\includegraphics[width=0.2\textwidth]{images/pascal/ResNet101-FCN/2007_000663.png} \\
				
				\includegraphics[width=0.2\textwidth]{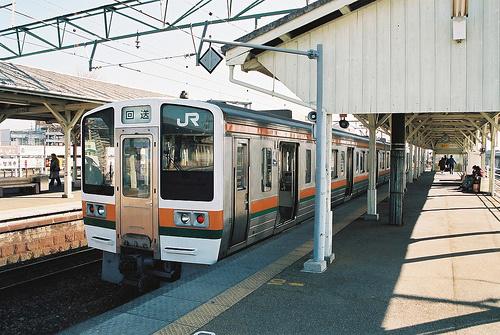}&
				\includegraphics[width=0.2\textwidth]{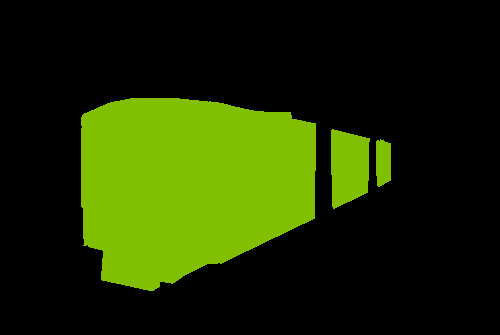}&
				\includegraphics[width=0.2\textwidth]{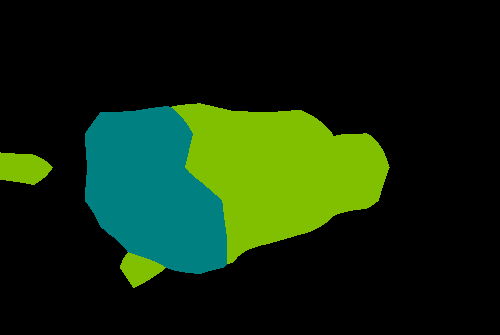}&
				\includegraphics[width=0.2\textwidth]{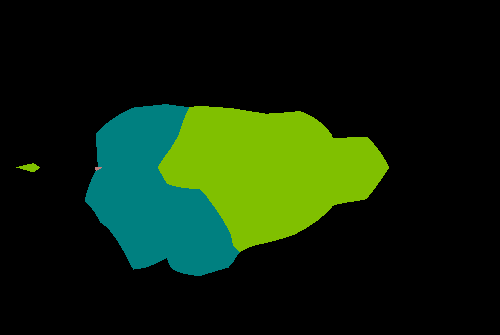}&
				\includegraphics[width=0.2\textwidth]{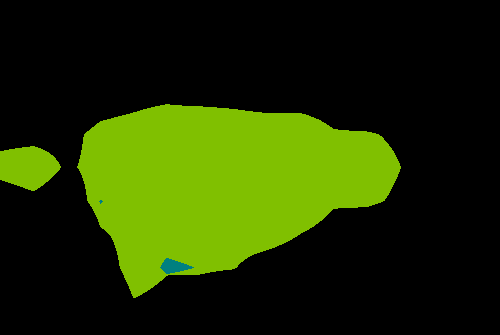} &
				\includegraphics[width=0.2\textwidth]{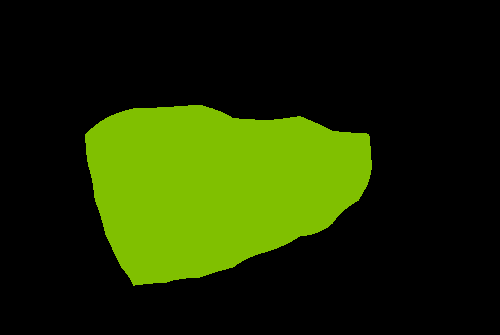}&				
				\includegraphics[width=0.2\textwidth]{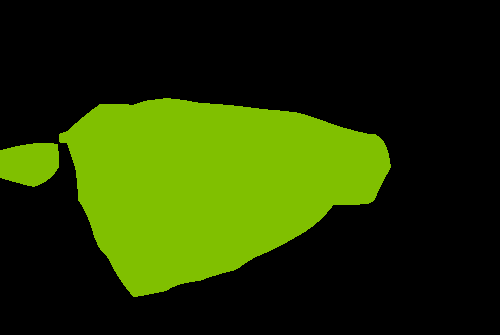}&
				\includegraphics[width=0.2\textwidth]{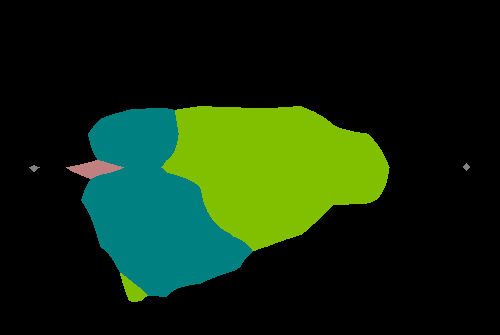} \\

				image & gt&ResNet50-FCN  & RIGNet $\mathbb{P}_u\ll6\gg$ & RIGNet $\mathbb{P}_u\ll6..5\gg$ & RIGNet $\mathbb{P}_u\ll6..4\gg$   & RIGNet $\mathbb{P}_u^{f}\ll6..1\gg$ & ResNet101-FCN \Tstrut

			\end{tabular}}
			\caption{Some samples of output quality after stage-wise addition of recurrent iterative gating modules. For each row, we show the input image, ground-truth, the vanilla ResNet-50 prediction, the predicted segmentation map of ResNet in a top$\rightarrow$down manner when iterative gating is included, and the output of vanilla ResNet101-FCN.}
			\label{fig:discuss}
		\end{center}
		\vspace{-0.5cm}
\end{figure*}

We use the same architectures and training procedures mentioned for evaluating performance on the COCO-Stuff dataset. Table~\ref{tab:cocostuff_val_fcnresnet} shows the quantitative comparison of our approach with respect to vanilla ResNet-FCN and DeepLabv2 based baselines. Our proposed \emph{ResNet50-RIGNet} achieves better mIoU (\textit{28.8}\% vs. \textit{24.3\%}) than the baseline. We further perform experiments on the COCO-Stuff 10k dataset with more sophisticated models (DeepLabV2-Res101). As shown in Table~\ref{tab:cocostuff_val_fcnresnet}, RIGNet consistently outperforms the baselines by a significant margin.
\begin{table}[h]
	\vspace{-0.3cm}
	\begin{center}
		\def\arraystretch{1.1}
		\resizebox{0.48\textwidth}{!}{
			\begin{tabular}{c|c|ccc}
				\specialrule{1.2pt}{1pt}{1pt}
				Methods& Parameters& pAcc & mAcc & mIoU  \\
				\specialrule{1.2pt}{1pt}{1pt}
				ResNet50&-& 57.2&35.2 &24.3\\
				ResNet50-RIGNet& $\mathbb{P}_u$$\ll6..4\gg$, $u_i=2$ &59.8&38.2& 26.6\\
				ResNet50-RIGNet& $\mathbb{P}_u$$\ll6..4\gg$, $u_i=6$ &61.4&40.0&  \cellcolor{Gray} \color{red}\textbf{28.8}\\
				\midrule
				ResNet101 &-&58.7&38.2&26.4\\
				ResNet101-RIGNet& $\mathbb{P}_u$$\ll6..4\gg$, $u_i=2$ &60.8&39.4& 28.0\\
				ResNet101-RIGNet& $\mathbb{P}_u$$\ll6..4\gg$, $u_i=6$&62.6&41.1& 29.5\\
				ResNet101-RIGNet& $\mathbb{P}_u^{f}\ll6..1\gg$, $u_i=6$ &62.3&41.6&  \cellcolor{Gray} \color{red}\textbf{29.9}\\
				\specialrule{1.2pt}{1pt}{1pt}
				DeepLabV2&-& 65.4&44.6&34.1 \\
				DeepLabV2-RIGNet& $\mathbb{P}_u$$\ll6..1\gg$, $u_i=2$ &66.1&46.6&  \cellcolor{Gray} \color{red}\textbf{35.0}\\
				\specialrule{1.2pt}{1pt}{1pt}
			\end{tabular}}
			\caption{Comparison of several ResNet based networks w/o and w/ RIG on COCO-Stuff 10K validation set.}
			\label{tab:cocostuff_val_fcnresnet}
		\end{center}
		\vspace{-0.5cm}
\end{table}

The superior performance achieved by RIGNet reveals that integrating recurrent iterative gating modules in the feed-forward network are very effective in capturing more contextual information for labeling complex scenes.
\section{Discussion}
Towards examining the practical grounds for recurrent iterative gating networks with top-down feedback, we aimed to verify a few specific hypotheses with our experiments. Firstly, the recurrent iterative gating mechanism can allow more parsimonious networks to outperform deeper architectures with careful selection of the gating structure. This is revealed to be the case for the RIGNet architecture, with clear evidence of ResNet50-RIGNet outperforming ResNet101-FCN with three iterations.

Secondly, our proposed RIGNet is more precise and semantically meaningful compared to the baselines with respect to qualitative results presented in Fig.~\ref{fig:val_pascal}. 
Fig.~\ref{fig:discuss} illustrates the impact of integrating a recurrent gating module. We can see that the recurrent iterative gating scheme progressively improves the details of a predicted segmentation map by recovering the missing spatial details which can be seen as coarse-to-fine refinement.

Moreover, the improvement in performance as a function of recurrent gating depth (RIG blocks) reveals that the RIGNet formulation of the feed-forward convolutional network improves the representational power of the model by incorporating semantic and relational context in a top-down manner. While there may exist alternatives for feedback gate design, we present an intuitive way of designing recurrent feedback gates. Given the demonstrable capability of the proposed mechanism, we expect that this work will inspire significant interest in exploring feedback based approaches within future work including an emphasis on top-down processing, gating, and iteration. 

\section{Conclusion}\label{sec:conclusion}
We have presented a network mechanism that involves \emph{Recurrent Iterative Gating} for semantic segmentation, showing it is a valuable contribution in its compatibility with virtually any feedforward neural network providing a general mechanism for boosting performance, or allowing for a simpler network. Our experimental results on two challenging datasets demonstrate that the proposed model performs significantly better than the baselines. We also found that feedback based approaches are capable of developing a meaningful coarse-to-fine representation similar to coarse-to-fine refinement based approaches without bringing complexity into the network architecture. As one specific example, ResNet-50 as a RIGNet is shown to outperform ResNet-101.

Results presented reveal the capability for correcting errors made in a single feedforward pass through Recurrent Iterative Gating as an exciting and important direction for future work, which even for deep networks, allows for stronger representational capacity. \\

\noindent \textbf{Acknowledgments:} The authors acknowledge financial support from the NSERC Canada Discovery Grants program, MGS funding, and the support of the NVIDIA Corporation GPU Grant Program.  

{\small
\bibliographystyle{ieee}
\bibliography{paper}
}

\end{document}